\documentclass[10pt]{article}

\usepackage{graphicx}
\usepackage{color}

\usepackage[authoryear]{natbib}

\usepackage{bm}
\usepackage{amsmath}
\usepackage{amsfonts}
\usepackage{amsthm}
\usepackage{amssymb}
\usepackage{mathtools}
\mathtoolsset{showonlyrefs=true}

\usepackage{url}
\usepackage{subcaption}
\usepackage{algorithm}
\usepackage{algorithmic}

\usepackage[hidelinks]{hyperref}

\usepackage{fancyvrb}  
\usepackage{xcolor}    

\DefineVerbatimEnvironment{CodeInput}
{Verbatim}
{
formatcom=\slshape,  
baselinestretch=1.2,                 
fontsize=\small                      
}

\DefineVerbatimEnvironment{CodeOutput}
{Verbatim}
{
formatcom=\color{black},             
frame=none,                          
baselinestretch=1.2,                
fontsize=\small                     
}

\newenvironment{CodeChunk}
{\vspace{0.5em}}  
{\vspace{0.5em}}

\usepackage{mcr}

\title{\texttt{si4onnx}: A Python package for Selective Inference in Deep Learning Models}

\date{\today}

\makeatletter
\def\@fnsymbol#1{\ensuremath{\ifcase#1\or
{1}\or
{2}\or
{3}\or
{4}\or
{5}\or
{\dagger}\or
\else\@ctrerr\fi}}
\makeatother

\author{
Teruyuki Katsuoka\footnotemark[1] ,
Tomohiro Shiraishi\thanks{Nagoya University} ,\\
Daiki Miwa\thanks{Nagoya Institute of Technology} ,
Shuichi Nishino\footnotemark[1] \thanks{RIKEN} ,\\
Ichiro Takeuchi \footnotemark[1] \footnotemark[3] \thanks{Corresponding author. e-mail: ichiro.takeuchi@mae.nagoya-u.ac.jp}
}
\begin{document}

\maketitle

\thispagestyle{empty}

\begin{abstract}
    \noindent
    In this paper, we introduce \texttt{si4onnx}, a package for performing selective inference on deep learning models.
Techniques such as CAM in XAI and reconstruction-based anomaly detection using VAE can be interpreted as methods for identifying significant regions within input images. However, the identified regions may not always carry meaningful significance. Therefore, evaluating the statistical significance of these regions represents a crucial challenge in establishing the reliability of AI systems.
\texttt{si4onnx} is a Python package that enables straightforward implementation of hypothesis testing with controlled type I error rates through selective inference. It is compatible with deep learning models constructed using common frameworks such as PyTorch and TensorFlow.
\end{abstract}

\newpage
\section{Introduction}
\label{sec:intro}
\begin{figure}[htbp]
  \centering
  \begin{minipage}{0.99\textwidth}
    \begin{tabular}{cc}
      \includegraphics[width=0.475\textwidth]{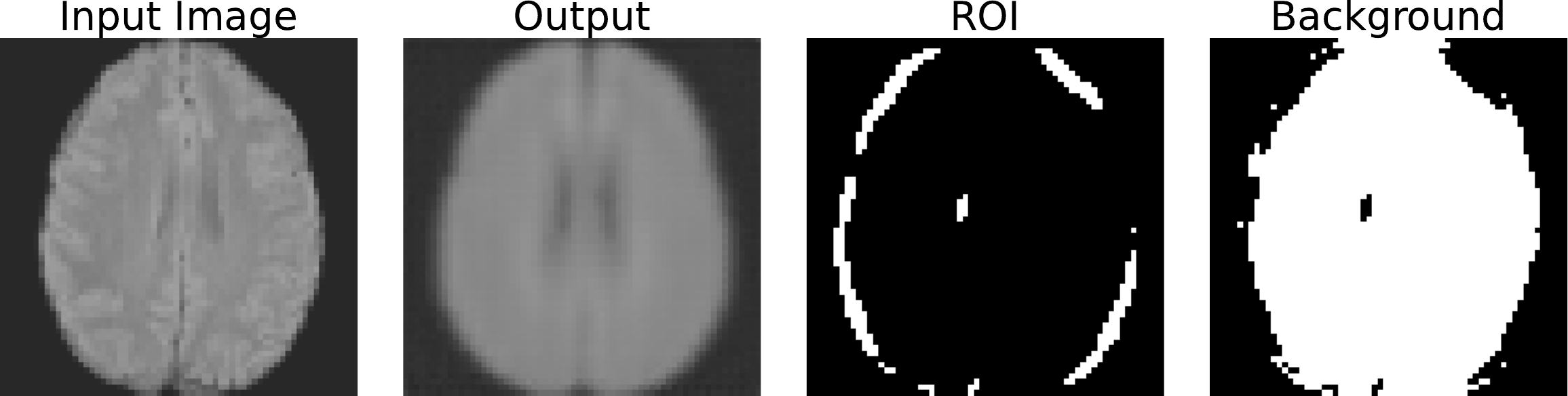} &
      \includegraphics[width=0.475\textwidth]{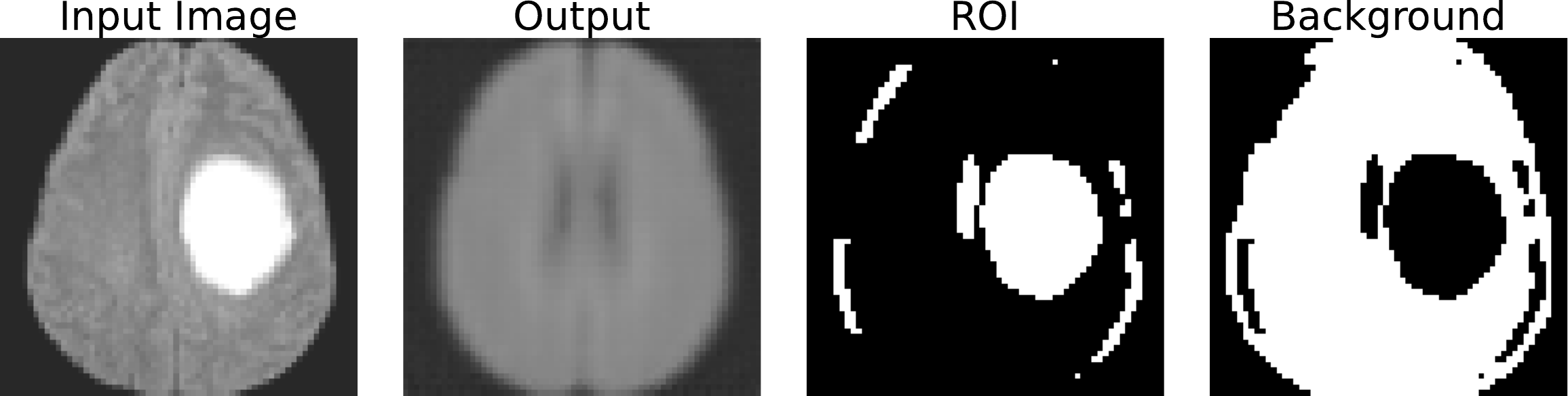} \\
      normal image &
      abnormal image \\
      naive $p$ = 0.000, selective $p$ = 0.185 &
      naive $p$ = 0.000, selective $p$ = 0.000 
    \end{tabular}
    \subcaption{Example 2: Anomaly Detection by Variational Auto-Encoder (VAE)}
    \label{subfig:vae}
    \vspace*{5.0mm}
  \end{minipage}
  \begin{minipage}{0.99\textwidth}
    \begin{tabular}{cc}
      \includegraphics[width=0.475\textwidth]{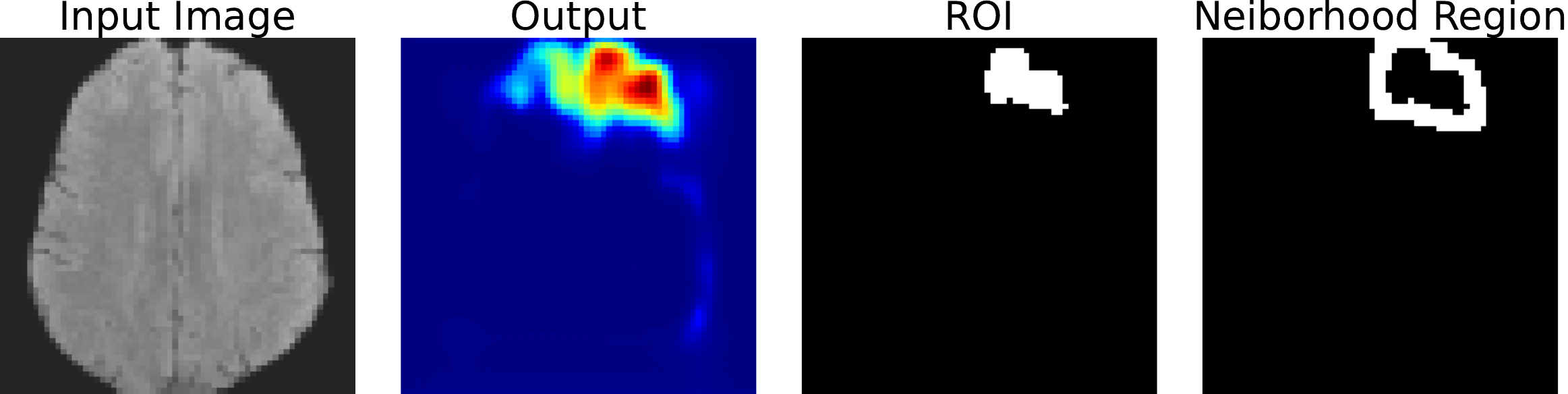} &
      \includegraphics[width=0.475\textwidth]{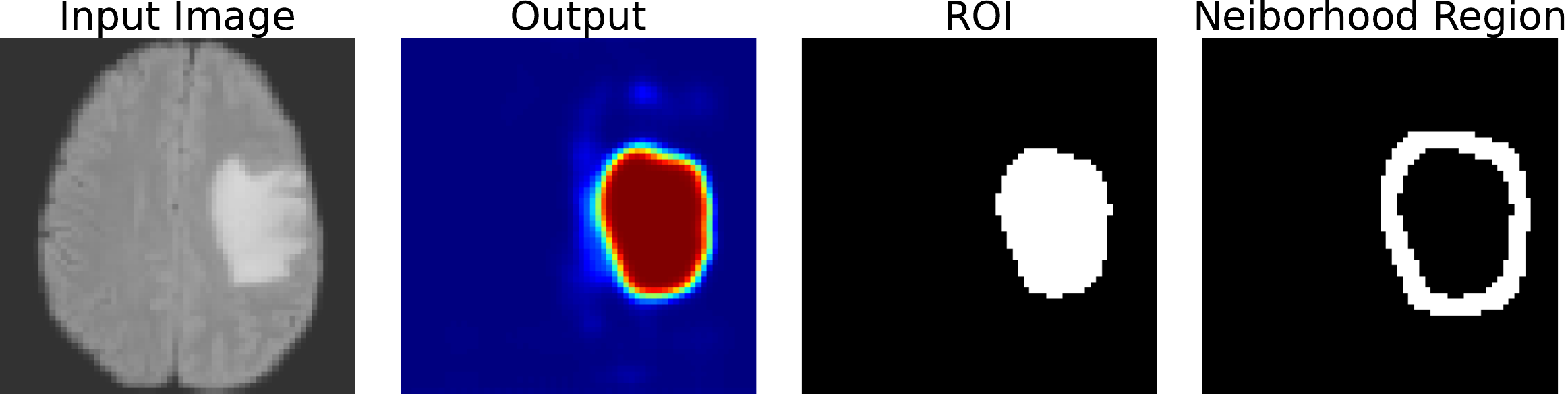} \\
      normal image &
      abnormal image \\
      naive $p$ = 0.000, selective $p$ = 0.709 &
      naive $p$ = 0.000, selective $p$ = 0.000 
    \end{tabular}
    \subcaption{Example 3: Segmentation by U-Net}
    \label{subfig:unet}
    \vspace*{5.0mm}
  \end{minipage}  
  \begin{minipage}{0.99\textwidth} 
    \begin{tabular}{cc}
      \includegraphics[width=0.475\textwidth]{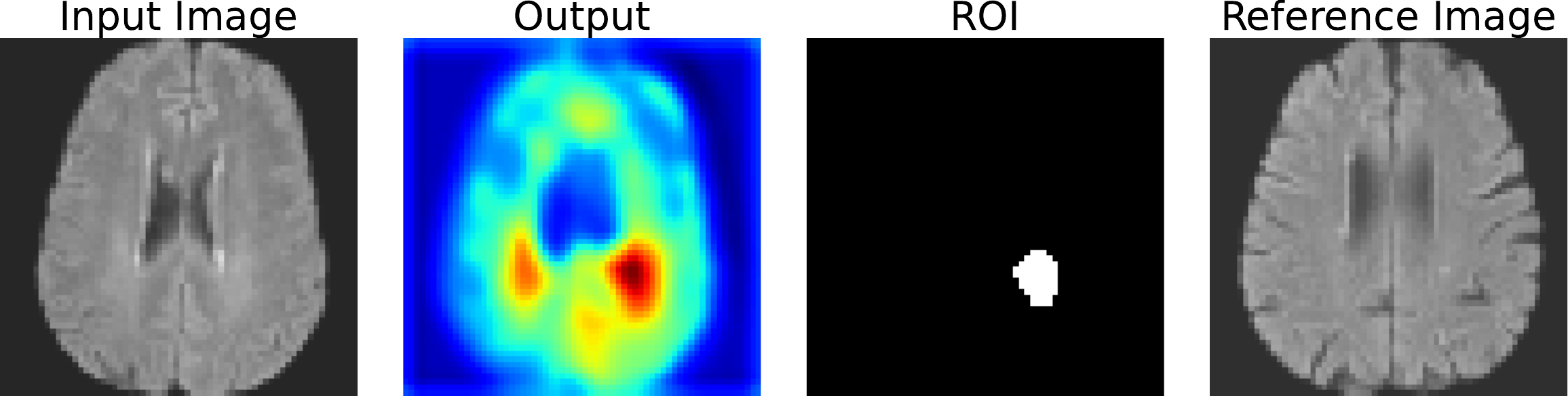} &
      \includegraphics[width=0.475\textwidth]{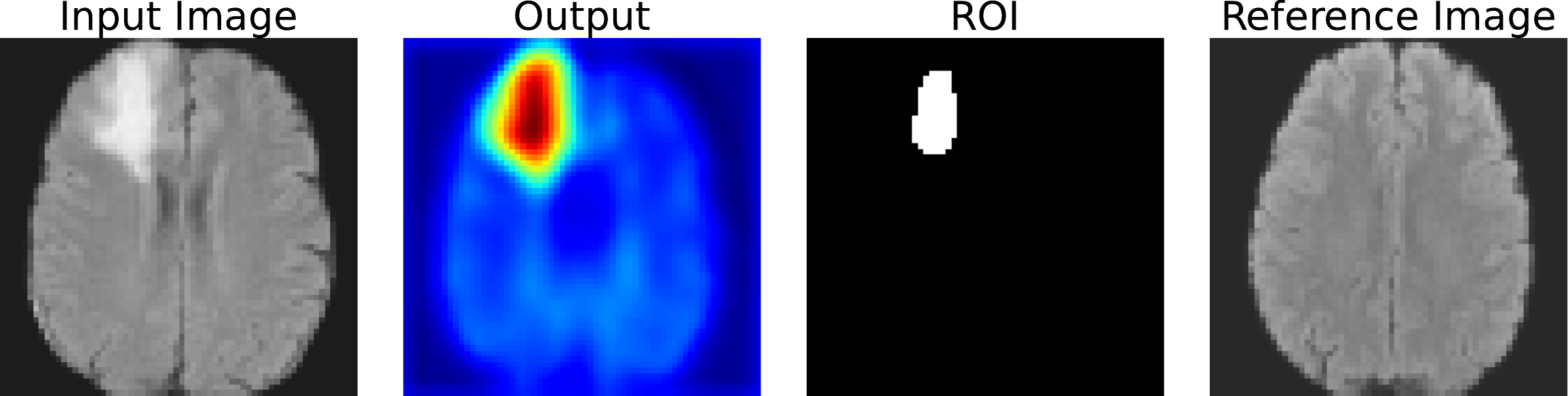} \\
      normal image &
      abnormal image \\
      naive $p$ = 0.002, selective $p$ = 0.915 &
      naive $p$ = 0.000, selective $p$ = 0.000
    \end{tabular}
    \subcaption{Example 1: Saliency Region by Class Activation Map (CAM)}
    \label{subfig:cam}
  \end{minipage}
\caption{
Three examples of statistical tests on ROIs detected by different types of DL models applied to BraTS 2023 (T2-FLAIR) MRI images \citep{Karargyris2023, labella2023asnrmiccaibraintumorsegmentation} are presented. Details of each problem setting and the experimental results can be found in Section~\ref{sec:problem_settings} and Section~\ref{sec:naive_selective_p_values} respectively. 
}
\label{fig:brain_mri}
\end{figure}

In deep learning (DL)-based image processing, detecting the \emph{region of interest (ROI)} enhances explainability and interpretability, providing a foundational step for various subsequent tasks.
For example, in DL-based image classification tasks, methods such as CAM~\citep{zhou2016learning} and GradCAM~\citep{selvaraju2017grad} are commonly used to identify ROIs that influence class assignments.
Furthermore, in image anomaly detection tasks, it is necessary to not only detect the presence of an anomaly in the image but also identify the anomalous region as an ROI~\citep{golan2018deep,bergmann2019mvtec}.
Segmentation and object detection for images can also be considered as a type of ROI detection~\citep{ronneberger2015u,long2015fully}.
However, there is no established method for quantifying the statistical significance of the detected ROIs, posing a barrier to using ROI detection results in high-stakes decision-making~\citep{adebayo2018sanity,dombrowski2019explanations}.
In this paper, we introduce a general software for quantifying the statistical significance of detected ROIs in the form of $p$-values, thereby allowing the quantification of false positive ROI detection probability.
Fig.~\ref{fig:brain_mri} provides examples of $p$-values calculated using the proposed software, evaluating the statistical significance of ROIs obtained by CAM, anomalous region detection, and segmentation, respectively.
Fig.~\ref{subfig:vae} shows a test image was analyzed using an anomaly detection method based on a Variational Auto-Encoder (VAE) as the generative model, with the detected region identified as the ROI (see Example A1 and Example B1 in Section~\ref{sec:problem_settings} for details).  
Fig.~\ref{subfig:unet} shows the results of statistical tests are shown for object regions obtained as ROIs by performing segmentation on test input images using a trained DL model called U-Net (see Example A2 and Example B2 in Section~\ref{sec:problem_settings} for details).
Fig.~\ref{subfig:cam} shows test image was input into a CNN trained to classify whether brain images are normal or abnormal, with regions where CAM values exceeded a certain threshold defined as ROIs (see Example A3 and Example B3 in Section~\ref{sec:problem_settings} for details).  
In each case (\subref{subfig:vae})-(\subref{subfig:cam}), the left part represents a healthy individual without a tumor, while the right part depicts a patient with a tumor, with both naive $p$-values and selective $p$-values provided for each case (see Section~\ref{sec:problem_settings} for details).  
In all three cases, the naive $p$-values for normal images are small, leading to the erroneous conclusion that the detected ROIs are statistically significant. In contrast, the selective $p$-values for normal images are large, correctly indicating that the detected ROIs are false detections.  
This paper introduces software for calculating selective $p$-values for DL models in various problem settings, including the cases (\subref{subfig:vae})-(\subref{subfig:cam}).

The problem of evaluating the statistical significance of detected ROIs within images poses technical challenges.
This is because the ROIs are detected from the image itself, and their statistical significance is also calculated using the same image. 
In the statistical community, this is often referred to as \emph{double-dipping}~\citep{kriegeskorte2009circular}, where the same data is used for both hypothesis selection and evaluation.
To address the bias arising from the double-dipping issue, we introduce a framework called \emph{selective inference (SI)}, which has recently gained attention~\citep{taylor2015statistical}.
The core idea of SI lies in conducting statistical inference based on the sampling distribution conditional on a hypothesis selection event.
Conditional $p$-values, calculated based on the sampling distribution conditional on a hypothesis selection event are referred to as selective $p$-values.
Selective $p$-values enable us to control type I error at the desired level, even in double-dipping situations where the hypothesis is selected based on the same data.
In order to compute selective $p$-values for ROIs detected by DL models, the challenge is how we can compute the sampling distribution of relevant test statistic condional on an event that an ROI is detected by the DL model.

Since the computational processes of DL models are complex, characterizing the selection event for ROIs seems intractable.
However, it has been shown within the SI research community that this is feasible for a certain class of DL models~\citep{duy2022quantifying,miwa2023valid,miwa2024statistical,katsuoka2024statisticaltestdiffusionmodelbased}.
Specifically, if the components of a DL model can be decomposed into piecewise linear functions, the hypothesis selection event can be expressed as a set of linear inequalities, allowing for the calculation of exact and valid $p$-values.
On the other hand, to realize this, it is necessary to derive and implement a large number of linear inequalities based on the specific structure of the DL model.
Given that the network structure of DL models varies depending  on the problem and data, deriving and implementing these inequalities for each DL model requires considerable effort.
To avoid this derivation and implementation cost, we have developed a software framework that leverages the modularity of DL models.
The proposed framework automatically derives the required linear inequalities for computing selective $p$-values without additional implementation costs.
Specifically, if the trained model is saved in the unified DL model format \emph{ONNX}~\footnote{DL models trained on many deep learning platforms, including PyTorch and TensorFlow, can be converted to the ONNX format.}, our software can provide selective $p$-values for the detected ROIs by the DL model.

\paragraph{Related Works}
SI was initially introduced as a method for quantifying statistical significance in feature selection for linear models~\citep{lee2014exact,lee2016exact,tibshirani2016exact}.
Subsequent research expanded SI to more complex feature selection methods~\citep{yang2016selective,suzumura2017selective,hyun2018exact,rugamer2020inference,das2021fast,garcia2023optimal,pirenne2024parametric}.
Moreover, it was recognized that SI could be applied to problems beyond feature selection, resulting in various extensions across different areas~\citep{chen2020valid,tsukurimichi2021conditional,tanizaki2020computing,duy2022quantifying,le2024cad,lee2015evaluating,gao2022selective,duy2020computing,jewell2022testing}.
Additionally, numerous approaches have been proposed to enhance the power of SI~\citep{tian2018selective,liu2018more,duy2022more,chen2023more}.
Research on SI for deep learning models includes applications to segmentation~\citep{duy2022quantifying}, saliency maps~\citep{miwa2023valid}, attentions~\citep{shiraishi2024statistical}, and anomaly detection~\citep{miwa2024statistical, katsuoka2024statisticaltestdiffusionmodelbased}.
However, these existing SI studies are specifically tailored to individual problems, models, and algorithms.
In other words, they derive selection events and compute conditional distributions for specific settings, requiring new derivations and implementations for SI whenever any aspect of the problem, model, or procedure changes.

Aside from the few SI studies mentioned earlier, to the best of our knowledge, no existing method effectively estimates the statistical significance of ROIs identified by DL models.
Alternative approaches, such as multiple testing correction, random permutation, or data splitting, could be considered for this purpose, but each poses challenges.
For instance, using Bonferroni correction as a multiple testing adjustment, the number of possible ROI combinations in an image with $n$ pixels is $2^n$, making direct application excessively conservative.
Another possible approach involves generating random images under the null hypothesis, such as pixel permutation.
However, when strong signals are concentrated in a small number of pixels, ROIs may still appear in the permuted images, making it difficult to accurately quantify their statistical significance.
Data splitting can often address the issue of double-dipping; however, in our case, where statistical inference is conducted on a single image, splitting the data (image) is not feasible.

\newpage
\section{Problem settings}
\label{sec:problem_settings}
This section outlines the problem settings addressed by the software proposed in this paper.

\subsection{Common setting}
We consider an image with $n$ pixels as an $n$-dimensional vector $\bm x \in \RR^n$ of pixel values.
We assume that an image $\bm x$ is a realization of an $n$-dimensional random vector
\begin{equation}
 \label{eq:data_vector}
 \bm X = \bm s + \bm \veps, 
\end{equation}
where
$\bm s \in \RR^n$
is the unknown $n$-dimensional vector of true signal values, 
$\bm \veps \in \RR^n$
is the Gaussian noise from
$\cN(\bm 0, \Sigma)$
with
$\Sigma$
the covariance matrix that is known or estimable from separate independent data. 
We interpret a trained deep learning model as an algorithm $\cA$ that maps the $n$-dimensional vector $\bm X$ to a set of pixels in the ROI, denoted by
\begin{equation}
 \label{eq:roi}
 \cA: \RR^n \ni \bm X \mapsto \cM_{\bm X} \in 2^{[n]},
\end{equation}
where $\cM_{\bm X}$ represents the set of pixels in the ROI, which is a subset of the $n$ pixels with $2^{[n]}$ indicating the power set of $[n] := \{1, 2, \ldots, n\}$.
For a class of DL models (which will be clarified in Section~\ref{sec:computing_selective_p_values}), our software is applicable to cases with any type of ROI identification methods.

As a general form of statistical test for quantifying the statistical significance of ROI, we consider the following null hypothesis ${\rm H}_0$ and alternative hypothesis ${\rm H}_1$:
\begin{align}
 \label{eq:test}
 {\rm H}_0:
 \bm \eta_{\cM_{\bm X}}^\top \bm s = 0
 \quad
 \text{v.s.,}
 \quad
 {\rm H}_1:
 \bm \eta_{\cM_{\bm X}}^\top \bm s \neq 0,
\end{align}
where $\bm \eta_{\cM_{\bm X}} \in \RR^n$ is a vector depending on $\cM_{\bm X}$, i.e., the ROI detected when the image $\bm X$ is input to the DL model.
For the statistical testing problem in \eqref{eq:test}, we can consider the following test statistic: 
\begin{equation}
 \label{eq:test_statistic}
 T(\bm X) = \bm \eta_{\cM_{\bm X}}^\top \bm X.
\end{equation}

To illustrate the practical application of this framework, specific examples of DL models and their ROI identification methods are presented in Section~\ref{subsec:roi_identification}, followed by concrete examples of test statistics in Section~\ref{subsec:examples_of_statistical_test}.

\subsection{Examples of ROI identification methods}
\label{subsec:roi_identification}
Here, we present three examples of ROI identification in images using DL models which can be managed by the proposed software.
\paragraph{Problem Example A1: ROI identification by anomaly detection.}
Consider a method for detecting anomalous regions using generative DL models.
When a tumor image is input into a deep generative model trained solely on normal images, the model produces a counterfactual image illustrating what the image would look like if it were normal.
By computing the difference between the generated counterfactual image and the original test image, the abnormal regions are quantified, and pixels with values exceeding a specified threshold are defined as the ROI.
Specifically, the algorithm $\cA: \bm{X} \mapsto \cM_{\bm{X}}$ is defined as a function that takes a test image $\bm{X}$ as input and returns the set of pixels corresponding to the anomalous regions detected by a generative DL model.

\paragraph{Problem Example A2: ROI idendification by segmentaion.}
Consider a DL model for segmentation trained on a dataset of images with annotated segmentation masks.
Here, a model called U-Net is used as the DL model suitable for segmentation.
Given a test image, the trained model can identify object region within the image, and the set of pixels corresponding to the object region identified by the segmentation model is interpreted as the ROI.
Namely, the algorithm $\cA: \bm{X} \mapsto \cM_{\bm{X}}$ is defined as a function that takes a test image $\bm{X}$ as input and returns the set of pixels in the object region as the segmentation result.

\paragraph{Problem Example A3: ROI Identification by CAM.}
Consider a DL model trained to classify whether brain images are normal or abnormal.
By incorporating specific components into a deep learning model for classification, it is possible to identify the local image regions that serve as the basis for the classification results, a technique known as saliency methods.
Here, we consider CAM, one of the representative saliency methods, and define the set of pixels with CAM values above a specific threshold as the ROI.
Namely, the algorithm $\cA: \bm X \mapsto \cM_{\bm X}$ is interpreted as a function that takes a test image $\bm X$ as input and returns the set of pixels with CAM values above the threshold for a trained DL model used for class classification.

\subsection{Examples of statistical test approaches for ROI's significance}
\label{subsec:examples_of_statistical_test}
Here, three examples discussed in Fig.~\ref{fig:brain_mri} are considered as testing approaches to quantify the statistical significance of ROI.

\paragraph{Test Example B1: Difference between ROI and background region.}
As the first example, we consider a simple approach for statistical testing of the ROI by comparing the pixel values in the ROI with those in the background region.  
Let $\cM_{\bm X}^c := [n] \setminus \cM_{\bm X}$ denote the set of pixels that are not part of the ROI.  
The mean difference between pixel values in the ROI and those in the background regions is expressed as:  
\begin{align*}
\frac{1}{|\cM_{\bm X}|} \sum_{i \in \cM_{\bm X}} X_i - \frac{1}{|\cM_{\bm X}^c|} \sum_{i \in \cM_{\bm X}^c} X_i. 
\end{align*}
By defining  
\begin{align*}
\bm \eta_{\cM_{\bm X}} :=
\frac{\bm 1_{\cM_{\bm X}}}{|\cM_{\bm X}|}
- 
\frac{\bm 1_{\cM^c_{\bm X}}}{|\cM^c_{\bm X}|},
\end{align*}
where $\bm 1_{\cC}$ denotes the $n$-dimensional indicator vector of set $\cC$, we can perform such a statistical test in Eqs.~(Eq.\eqref{eq:test} and Eq.\eqref{eq:test_statistic}).

\paragraph{Test Example B2: Difference between ROI and its neighborhood region.}
The last example involves comparing the pixel values in the identified ROI with those in the \emph{neighborhood} region of the ROI.  
The neighborhood regions can be arbitrarily defined, for instance, as a set of pixels located within a certain distance from the ROI.  
Let the set of pixels in the neighborhood region be denoted as $\cM^{\rm nbd}_{\bm X}$.  
The mean difference between the pixel values in the ROI and those in the neighborhood region is given by:  
\begin{align*}
\frac{1}{|\cM_{\bm X}|} \sum_{i \in \cM_{\bm X}} X_i - \frac{1}{|\cM_{\bm X}^{\rm nbd}|} \sum_{i \in \cM_{\bm X}^{\rm nbd}} X_i.
\end{align*}
By defining  
\begin{align*}
\bm \eta_{\cM_{\bm X}} :=
\frac{\bm 1_{\cM_{\bm X}}}{|\cM_{\bm X}|}
- 
\frac{\bm 1_{\cM^{\rm nbd}_{\bm X}}}{|\cM^{\rm nbd}_{\bm X}|},
\end{align*}
in Eqs.~\eqref{eq:test} and \eqref{eq:test_statistic}, we can conduct such a statistical test.

\paragraph{Test Example B3: Difference between ROI and corresponding region in a reference image.}
Yet another approach for testing the statistical significance of the identified ROI is to compare it with the same corresponding in normal reference image. 
In this case, we need to slightly change the notation from Eq.\eqref{eq:data_vector}.
Let us denote a test image and a reference image as
\begin{align*}
 \bm X^{\rm tst} = \bm s^{\rm tst} + \bm \veps^{\rm tst}, \\
 \bm X^{\rm ref} = \bm s^{\rm ref} + \bm \veps^{\rm ref}, 
\end{align*}
where
are the unknown true signals,
while 
$\bm \veps^{\rm tst} \sim N(\bm 0, \Sigma^{\rm tst})$
and 
$\bm \veps^{\rm ref} \sim N(\bm 0, \Sigma^{\rm tst})$
are the noise vectors with known or estimable covariance matrices
$\Sigma^{\rm tst}$
and 
$\Sigma^{\rm ref}$,
respectively. 
The mean difference of pixel values in the ROI between the test and the reference images is represented as
\begin{align*}
 \frac{1}{|\cM_{\bm X^{\rm tst}}|} \sum_{i \in \cM_{\bm X^{\rm tst}}} X_i^{\rm tst}
 - 
 \frac{1}{|\cM_{\bm X^{\rm tst}}|} \sum_{i \in \cM_{\bm X^{\rm tst}}} X_i^{\rm ref}. 
\end{align*}
With a slight abuse of notations, if we write
\begin{align*}
 \bm X := \mtx{c}{
 \bm X^{\rm tst} \\
 \bm X^{\rm ref} 
 },
 \bm s := \mtx{c}{
 \bm s^{\rm tst} \\
 \bm s^{\rm ref} 
 },
 \Sigma = \mtx{cc}{
  \Sigma^{\rm tst} & \bm 0 \\
  \bm 0 & \Sigma^{\rm ref}
  },
\end{align*}
and define 
\begin{align*}
 \bm \eta_{\cM_{\bm X}} := 
 \frac{1}{|\cM_{\bm X^{\rm tst}}|} 
 \begin{pmatrix}
  \bm 1_{\cM_{\bm X^{\rm tst}}} \\
  -\bm 1_{\cM_{\bm X^{\rm tst}}}
  \end{pmatrix},
\end{align*}
in Eqs.~\eqref{eq:test} and \eqref{eq:test_statistic}, we can perform such a statistical test.

\subsection{Applicable class of problem settings}
The proposed software contains presets for the above three ROI identification methods and the above three testing approaches, allowing any combination of these to be applied to a class of DL models described in Section~\ref{sec:computing_selective_p_values} (The examples in Fig.~\ref{fig:brain_mri} (\subref{subfig:vae}), (\subref{subfig:unet}), and (\subref{subfig:cam}) use Example A1, A2, and A3 for ROI identification, and B1, B2, and B3 for the testing approaches, respectively).
Additionally, users can specify their own ROI identification methods or testing approaches in the form of Eq.\eqref{eq:test_statistic}, allowing selective $p$-values to be computed without any additional implementation cost.
Furthermore, it is worth noting that any linear filter, such as difference filters, edge filters, can also be applied to test images. 

\section{Naive $p$-values and Selective $p$-values}
\label{sec:naive_selective_p_values}
\subsection{Naive $p$-values}
%
%
For now, let us consider an unrealistic situation where the ROI is determined independently of the image, that is, $\bm \eta_{\cM_{\bm X}}$ does not depend on $\bm X$, where 
\begin{equation}
  T(\bm X)
  \sim
  \cN\left(
  \bm 0,
  \bm \eta_{\cM_{\bm X}}^\top
  \Sigma
  \bm \eta_{\cM_{\bm X}}
  \right)
  ~
  \text{if $\bm \eta_{\cM_{\bm X}}$ {\it dose not depend on} $\bm X$}.
  \label{eq:naive_sampling_dist}
\end{equation}
We call the $p$-values obtained based on the sampling distribution in Eq.\eqref{eq:naive_sampling_dist} as \emph{naive $p$-values}. 
%
%
%
The naive $p$-value is valid when the ROI is determined independently of the image $\bm X$. 
However, in reality, since the ROI is determined based on the image $\bm X$, conducting statistical testing based on the naive $p$-value introduces bias and fails to properly evaluate the statistical significance of the ROI.
\subsection{Selective $p$-values}
\label{subsec:selective_inference}
%
The SI framework enables us to eliminate bias and compute valid $p$-values.
%
%
Within the SI framework, we consider the conditional distribution of $T(\bm{X})$ on the event that the same ROI as the observed one is obtained, i.e., 
\begin{equation}
  \label{eq:conditional_distribution}
  T(\bm{X}) \mid \left\{\mathcal{M}_{\bm{X}} = \mathcal{M}_{\bm{x}}\right\}, 
\end{equation}
where, remember that, $\cM_{\bm X}$ is random, while $\cM_{\bm x}$ is the observed ROI. 
Under this conditioning, the ROI $\cM_{\bm X}$ is fixed, making the test statistic $T(\bm{X})$ linear with respect to $\bm X$.
However, since this conditioning is imposed on a subspace of an $n$-dimensional space, computing this conditional distribution is computationally intractable.
To address this issue, it is common in SI literature to introduce additional conditioning through a nuisance parameter
\begin{align*}
 \mathcal{Q}_{\bm{X}} = \left( I_n - \frac{\Sigma \bm{\eta}_{\mathcal{M}_{\bm{X}}} \bm{\eta}_{\mathcal{M}_{\bm{X}}}^{\top} } {\bm{\eta}^{\top}_{\mathcal{M}_{\bm{X}}} \Sigma \bm{\eta}_{\mathcal{M}_{\bm{X}}}} \right) \bm{X},
\end{align*}
which is independent of the test statistic $T(\bm X)$. 
This common SI technique reduces the conditional space from $n$-dimensions to $1$-dimension.
Specifically, we consider the conditional distribution 
\begin{equation}
  \label{eq:conditional_tx}
  T(\bm{X}) \mid \left\{\mathcal{M}_{\bm{X}} = \mathcal{M}_{\bm{x}}, \mathcal{Q}_{\bm{X}} = \mathcal{Q}_{\bm{x}}\right\}
\end{equation}
for computing valid $p$-values.
The so-called \emph{selective $p$-value} is computed as 
\begin{equation}
  p_{\mathrm{selective}} = \mathbb{P}_{\mathrm{H}_0} \left( |T(\bm{X})| > |T(\bm{x})| \mid \bm{X} \in \mathcal{X} \right),
  \label{eq:p_selective_n_dim}
\end{equation}
where
%
\begin{equation}
  \mathcal{X} = \left\{ \bm{X} \in \mathbb{R}^n \mid \mathcal{M}_{\bm{X}} = \mathcal{M}_{\bm{x}}, \mathcal{Q}_{\bm{X}} = \mathcal{Q}_{\bm{x}} \right\}.
\end{equation}
Since the conditional data space is 1-dimensional space, we can represent the vector $\bm x$ using a scalar variable $z \in \mathbb{R}$ and then the conditional space $\mathcal{X}$ is parametrized by $z$ as 
\begin{equation}
  \label{eq:XandZ}
  \mathcal{X} = \left\{ \bm{X}(z) \in \mathbb{R}^n \mid \bm{X}(z) = \bm{a} + \bm{b}z, z \in \mathcal{Z} \right\},
\end{equation}
where 
\begin{equation}
  \label{eq:a_and_b}
  \bm{a} = \mathcal{Q}_{\bm x}, \;
  \bm{b} = \frac{\Sigma \bm{\eta}_{\mathcal{M}_{\bm x}}}
            {\bm{\eta}^{\top}_{\mathcal{M}_{\bm x}} 
              \Sigma
              \bm{\eta}_{\mathcal{M}_{\bm x}}},
\end{equation}
and 
\begin{equation}
  \label{eq:Z}
  \mathcal{Z} = \left\{z \in \mathbb{R} \mid \mathcal{M}_{\bm{X}(z)} = \mathcal{M}_{\bm x} \right\}.
\end{equation}

Let $Z \in \mathbb{R}$ be a random variable and $z_{\bm x}$ be its observation obtained from a test image $\bm x$, where $\bm X = \bm{a} + \bm{b}Z$ and $\bm x = \bm{a} + \bm{b}z_{\bm x}$.
The selective $p$-value can then be defined as:
\begin{align}
  \label{eq:selective_p_value_on_z}
  p_{\mathrm{selective}} = \mathbb{P}_{\mathrm{H}_0} \left( |Z| > |z_{\bm x}| \mid Z \in \mathcal{Z} \right).
\end{align}
If conditions are not taken into account, $Z$ follows the normal distribution $\mathcal{N}(\bm{0}, \bm{\eta}_{\mathcal{M}_{\bm{X}}}^\top \Sigma \bm{\eta}_{\mathcal{M}_{\bm{X}}})$ under the null hypothesis $\mathrm{H}_0$.
Consequently, the conditioned random variable $Z \in \mathcal{Z}$ follows a truncated normal distribution $\mathcal{TN}$. 
Therefore, computing $p_{\mathrm{selective}}$ requires determining the truncated intervals $\mathcal{Z}$.

\newpage
\section{Computing selective $p$-values}
\label{sec:computing_selective_p_values}
In this section, we present a method for calculating selective $p$-values for a class of DL models using parametric programming.
The core concept behind this parametric programming-based approach has already been explored in previous studies~\citet{duy2022quantifying,miwa2023valid,miwa2024statistical,katsuoka2024statisticaltestdiffusionmodelbased}. 
As mentioned in Section~\ref{sec:intro}, a key limitation of these studies is the need for labor-intensive derivation and implementation of selection events tailored to each DL model and problem setting.
In contrast, the software introduced in this paper automates the derivation of these selection events by simply loading a trained NN in ONNX format, thereby enabling the computation of selective $p$-values without additional implementation efforts.

\subsection{Piecewise-linear DL models}
\label{sec:over-conditioning}
The computation of truncated intervals in SI for deep learning models was proposed by \cite{duy2022quantifying} and further developed by \cite{miwa2023valid}.
They named this method \textit{Auto-Conditioning}.
This method utilizes the piecewise linearity of deep learning models to express the selection event as a set of linear inequalities.
A piecewise-linear algorithm $\cS(\bm{X})$ can be expressed as a set of linear functions in a polytope $\mathcal{P}_k$ in an $n$-dimensional space.
\begin{equation}
  \label{eq:A}
    \cS(\bm{X}) = \begin{cases}
      \bm{A}_2 + \bm{B}_1\bm{X} & \text{if } \bm{X} \in \mathcal{P}_1 ,\\
      \bm{A}_2 + \bm{B}_2\bm{X} & \text{if } \bm{X} \in \mathcal{P}_2 ,\\
      \qquad \vdots \\
      \bm{A}_K + \bm{B}_K\bm{X} & \text{if } \bm{X} \in \mathcal{P}_K ,
    \end{cases}
    \;\;\; \text{where} \;\; \mathcal{P}_k = \left\{\bm{C}_k \bm{X} \leq \bm{D}_k\right\}, k \in [K],
\end{equation}
where $K$ is the number of linear functions in the piecewise-linear function $\cS$.
The matrices $\bm{C}_k$ and $\bm{D}_k$ represent coefficient and constant matrices, respectively.
Similarly, $\bm{A}_k$ and $\bm{B}_k$ are matrices representing the intercept and coefficient of $\cS$ in the $k$-th polytope $\mathcal{P}_k$.
Following the discussion in \ref{subsec:selective_inference}, since the input image $\bm{X}(z)$ is parametrized by a 1-dimensional line $z$, it can be rewritten as follows:
\begin{equation}
  \label{eq:A_i}
  \begin{gathered}
      \cS_i(\bm{X}(z))
      = \begin{cases}
        a^{\cS_i}_1 + b^{\cS_i}_1 z & \text{if } z \in [L^{\cS_i}_1, U^{\cS_i}_1 ] ,\\
        a^{\cS_i}_2 + b^{\cS_i}_2 z & \text{if } z \in [L^{\cS_i}_2, U^{\cS_i}_2 ] ,\\
        \qquad \vdots \\
        a^{\cS_i}_{K(\cS_i)} + b^{\cS_i}_{K(\cS_i)} z & \text{if } z \in [L^{\cS_i}_{K(\cS_i)}, U^{\cS_i}_{K(\cS_i)} ],
      \end{cases}
  \end{gathered}
\end{equation}
where $K(\cS_i)$ is the number of linear functions in the piecewise-linear function $\cS_i$.
For $k \in [K(\cS_i)]$, $a^{\cS_i}_k$ and $b^{\cS_i}_k$ denote the intercept and coefficient scalar of $\cS_i$ in the $k$-th interval; these should be distinguished from $\bm a$ and $\bm b$ in \eqref{eq:a_and_b}.
\paragraph{Auto-Conditioning.}
\label{para:auto-conditioning}
Auto-Conditioning identifies the intervals $[L^{\cS_i}_k, U^{\cS_i}_k]$ ($k \in [K(\cS_i)]$) where each piecewise-linear function in \eqref{eq:A_i} exhibits linear behavior.
Since deep learning models accumulate linear functions through piecewise-linear operations such as ReLU and Max Pooling, we must iteratively identify increasingly constrained intervals after applying each layer.
To illustrate this process, consider a deep learning model with $N$ layers composed of convolution and ReLU operations.
Let us examine the $i$-th element at the $j$-th layer ($j \in [N]$) of the input vector parameterized by the scalar $z$.
Suppose the intermediate representation $a^{\cS_{i,j}} + b^{\cS_{i,j}} z$ is obtained through a linear transformation, and we have already determined the interval $z \in [l^{\cS_{i,j}}, u^{\cS_{i,j}}]$ where linear operations hold up to the $j$-th layer.
When applying ReLU to the $j$-th intermediate representation, the transformation takes the form:
\begin{equation}
  \begin{aligned}
  a^{\cS_{i,j+1}} + b^{\cS_{i,j+1}} z 
  &=\text{ReLU} (a^{\cS_{i,j}} + b^{\cS_{i,j}} z) \\
  &= 
  \begin{cases}
    a^{\cS_{i,j}} + b^{\cS_{i,j}} z & \text{if } a^{\cS_{i,j}} + b^{\cS_{i,j}} z > 0,\\
    0 & \text{otherwise}.
  \end{cases}
\end{aligned}
\end{equation}
This ReLU operation transforms the linear function into a piecewise-linear function with two distinct regions.
Let $\phi_{i,j}(z) := a^{\cS_{i,j}} + b^{\cS_{i,j}} z$ represent the intermediate representation at the $j$-th layer. Then, the interval $[l^{\cS_{i,j+1}}, u^{\cS_{i,j+1}}]$ is given by:
\begin{equation}
  [l^{\cS_{i, j+1}}, u^{\cS_{i, j+1}}] =  \\
  \begin{cases}
    \left[ \max \left( l^{\cS_{i,j}}, -\frac{a^{\cS_{i,j}}}{b^{\cS_{i,j}}}   \right)
        , u^{\cS_{i,j}} 
      \right]  
      & \text{if } \text{sign}(b^{\cS_{i,j}}) = \text{sign}(\phi_{i,j}(z)), \\
    \left[ l^{\cS_{i,j}}, 
        \min \left(u^{\cS_{i,j}}, -\frac{a^{\cS_{i,j}}}{b^{\cS_{i,j}}}   \right)
      \right]  
      & \text{otherwise}.
  \end{cases}
\end{equation}
Applying this procedure sequentially through all layers yields the final interval for the $N$-th layer, denoted as:
$[L^{\cS_i}_k, U^{\cS_i}_k] := [l^{\cS_{i, N}}, u^{\cS_{i, N}}]$.

\subsection{Parametric programming-based SI}

\paragraph{Over-conditioning.} 
\label{para:over-conditioning}
To determine the interval where the same ROI $\cM_{\bm X}$ can be obtained, we consider a threshold $\tau$ that defines the ROI selection.
For \eqref{eq:A_i}, let $\psi_{i,k}(z) := a^{\cS_i}_k + b^{\cS_i}_k z$ represent the linear function in the $k$-th interval $[L^{\cS_i}_k, U^{\cS_i}_k]$ for each $k \in [K(\cS_i)]$. We then determine the interval $\cZ^{\rm{oc}}$ where the condition $\psi_{i,k}(z) > \tau$ selects the same set of pixels.
For each pixel $i$, we first compute the interval $[L^i_z, U^i_z]$ that maintains the selection status:
\begin{equation}
\label{eq:interval_z}
\begin{gathered}
[L^i_z, U^i_z] \coloneqq
\begin{cases}
\left[ \max \left( L^{\cS_i}_k, \frac{\tau - a^{\cS_i}_k}{b^{\cS_i}_k} \right), U^{\cS_i}_k\right] 
  &\text{if } \text{sign}(b^{\cS_i}_k) = \text{sign}(\psi_{i,k}(z) - \tau), \\
\left[ L^{\cS_i}_k, \min \left( U^{\cS_i}_k, \frac{\tau - a^{\cS_i}_k}{b^{\cS_i}_k} \right) \right] 
  &\text{otherwise}.
\end{cases}
\end{gathered}
\end{equation}
The interval $\mathcal{Z}^{\rm{oc}}$ is obtained by intersecting these intervals across all pixels:
\begin{equation}
\label{eq:oc}
\mathcal{Z}^{\rm{oc}} (\bm{a} + \bm{b}z) = \bigcap_{i \in [n]}  \left[ L^i_z, U^i_z \right].
\end{equation}
Consequently, the $p$-value for over-conditioning can be computed as:
\begin{equation}
\label{eq:oc_p_value}
p_{\mathrm{oc}} = \mathbb{P}_{\mathrm{H}_0} \left( |Z| > |z_{\bm{x}}| \mid Z \in \mathcal{Z}^{\mathrm{oc}}(\bm{a}+\bm{b}z_{\bm{x}}) \right).
\end{equation}
Although this approach is computationally tractable, over-conditioning through intermediate layers causes excessive conditioning, resulting in reduced statistical power.
To address this limitation, we introduce a technique called \textit{Parametric Programming}~\citep{duy2022more} which eliminates the excessive conditioning.
\paragraph{Parametric Programming.}
%
To compute $p$-value in \eqref{eq:selective_p_value_on_z}, parametric programming 
comprehensively explores intervals along the 1-dimensional line $z$.
Given $z_{\bm x}$, we identify all truncated intervals $\mathcal{Z}$ that yield the same salient region $\mathcal{M}_{\bm{X}(z)}(= \mathcal{M}_{\bm{X}_{\bm{a}+\bm{b}z}})$ as $\mathcal{M}_{\bm x}$.
These intervals can be expressed using $\mathcal{Z}^{\mathrm{oc}}$ as:
\begin{equation}
 \label{eq:z_by_z_oc}
 \mathcal{Z} = \bigcup_{z \in \mathbb{R} \mid \mathcal{M}_{\bm{X}(z)} = \mathcal{M}_{\bm{x}}} \mathcal{Z}^{\mathrm{oc}} (\bm{a} + \bm{b}z).
\end{equation}
While the number of $\mathcal{Z}^{\mathrm{oc}}$ is finite due to the finite number of polytopes, exploring all polytopes is computationally intractable as their number grows exponentially with the depth and width of the network.
Fortunately, \cite{Shiraishi2024} showed that we can achieve sufficient accuracy by restricting the search range $z$ with appropriate bounds $z_{\min}$ and $z_{\max}$.
The detailed procedure for computing $p$-values by parametric programming is shown in Algorithm \ref{alg:parametric-si}.
\begin{algorithm}[H]
  \caption{Selective $p$-value Computation by Parametric Programming}
  \label{alg:parametric-si}
  \begin{algorithmic}[1]
  \REQUIRE $\bm{x}, z_{\min}, z_{\max}$ and $z_{\bm{x}}:= T(\bm{x})$
  \STATE $\mathcal{Z} \leftarrow \emptyset$
  \STATE Obtain $\mathcal{M}_{\bm{x}}$ by \eqref{eq:roi}
  \STATE Compute $\bm{a}$, $\bm{b}$ by \eqref{eq:a_and_b}
  \STATE $z \leftarrow z_\mathrm{min}$
  \WHILE {$z < z_{\max}$}
  \STATE Compute $\mathcal{Z}^{\mathrm{oc}}(\bm{a} + \bm{b} z)$ and $\mathcal{M}_{X(z)}$ by \eqref{eq:oc} for $z$
  \IF {$\mathcal{M}_{\bm{X}(z)} = \mathcal{M}_{\bm{x}}$}
  \STATE $\mathcal{Z} \leftarrow \mathcal{Z} \cup \mathcal{Z}^{\mathrm{oc}}(\bm{a} + \bm{b} z)$
  \ENDIF
  \STATE $z \leftarrow \max{\mathcal{Z}^{\mathrm{oc}}(\bm{a} + \bm{b} z)} + \epsilon$, where $\epsilon$ is small positive number.
  \ENDWHILE
  \STATE  $p_{\mathrm{selective}} = \mathbb{P}_{\mathrm{H}_0} \left( |Z| > |z_{\bm{x}}| \mid Z \in \mathcal{Z} \right)$
  \ENSURE $p_{\mathrm{selective}}$
  \end{algorithmic}
\end{algorithm}

\newpage
\section{The si4onnx Package}
\label{sec:si4onnx_package}
The \texttt{si4onnx} package provides tools for assessing the statistical significance of ROIs detected by deep learning models through SI-based $p$-value computation.
The package is available at \url{https://github.com/Takeuchi-Lab-SI-Group/si4onnx} and can be installed as follows:
\begin{CodeChunk}
 \begin{CodeInput}
 >>> pip install si4onnx
 \end{CodeInput}
\end{CodeChunk}
The \texttt{si4onnx} requires models to be saved in the ONNX format `\texttt{*.onnx}' and supports models with standard linear and piecewise-linear operations such as convolution and ReLU.
A comprehensive list of supported layers is available in the documentation at \url{https://takeuchi-lab-si-group.github.io/si4onnx/}.
Furthermore, the \texttt{si4onnx} supports models with multiple inputs and outputs, allowing users to specify the relevant input-output pairs for hypothesis testing.
\subsection{Implementation and Usage}
\label{subsec:implementation}
We demonstrates $p$-value computation using synthetic data generated from a Gaussian distribution $\mathcal{N}(\bm{0}, I)$.
The \texttt{SyntheticDataset} class generates synthetic data following a Gaussian distribution.
Users can specify the mean and standard deviation using the optional arguments \texttt{loc} and \texttt{scale}; by default, data follows the standard normal distribution.
For testing purposes, the \texttt{local\_signal} argument adds a signal to a random region of the image.
For 2D images, this signal is added to a randomly-positioned square region, with a size defined in pixels by the \texttt{local\_size} argument.
By default, this size is set to $1/3$ of the smaller dimension (width or height) of the image.
The following code generates a synthetic image of size \texttt{(1, 16, 16)} without additional signals:
\begin{CodeChunk}
 \begin{CodeInput}
 >>> from torch.utils.data import DataLoader
 >>> data = si4onnx.data.SyntheticDataset(
           n_samples=1, shape=(1, 16, 16), local_signal=0, seed=0
     )
 >>> dataloader = DataLoader(data, batch_size=1)
 >>> x, mask, label = next(iter(dataloader))
 \end{CodeInput}
\end{CodeChunk}
Computing $p$-values involves three steps:
\begin{enumerate}
 \item Load a deep learning model from an `\texttt{.onnx}' file.
 \item Initialize an SI-executable model using the \texttt{load} function, specifying the desired hypothesis.
 \item Compute the $p$-value via the \texttt{inference} method, providing the observed image ${\bm{x}}$ and variance $\Sigma$.
\end{enumerate}
The following example demonstrates selective $p$-value computation using CAM on synthetic images:
\begin{CodeChunk}
  \begin{CodeInput}
  >>> import onnx
  >>> import si4onnx
  >>> onnx_model = onnx.load("cam_16.onnx")
  >>> si_model = si4onnx.load(
  ...     model=onnx_model,
  ...     hypothesis=si4onnx.hypothesis.BackMeanDiff(
  ...         threshold=0.9,
  ...         use_norm=True,
  ...     ),
  ... )
  >>> result = si_model.inference(x, var=1.0)
  >>> print(result.p_value)
  \end{CodeInput}
\end{CodeChunk}
\begin{CodeChunk}
  \begin{CodeInput}
  0.4288679844549881
  \end{CodeInput}
\end{CodeChunk}
The \texttt{load} function accepts the following arguments:
\begin{itemize}
 \item \texttt{model}: The ONNX model loaded using \texttt{onnx.load}.
 \item \texttt{hypothesis}: 
 An instance of the \texttt{PresetHypothesis} class, which provides three preset hypotheses: \texttt{BackMeanDiff}, \texttt{NeighborMeanDiff}, and \texttt{ReferenceMeanDiff}. 
 Details of these hypotheses are described in Section~\ref{subsec:preset_hypotheses}.
 \item \texttt{mask}:
 A binary mask indicating the region not contains of ROI $\cM_{\bm X}$ and other regions (e.g., background regions, neighborhood regions).
 \item \texttt{seed}: A optional random seed for random number generators (e.g., VAEs). Defaults to \texttt{None}.
 \item \texttt{memoization}:
 A boolean flag to enable memoization of interval calculations.
 This speeds up computation and reduces memory usage.
 Defaults to \texttt{True}.
\end{itemize}
The \texttt{InferenceResult} class provides the following attributes:
\begin{itemize}
 \item \texttt{p\_value}: The $p$-value computed by the SI framework.
 \item \texttt{naive\_p\_value()}: The naive $p$-value.
 \item \texttt{bonferroni\_p\_value()}: Bonferroni-corrected $p$-value.
 \item \texttt{output}: Deep learning model outputs.
 \item \texttt{score\_map}: The score map $\cS(\bm{x})$.
 \item \texttt{roi}: The ROI $\cM_{\bm{x}}$.
 \item \texttt{non\_roi}: The complement set of ROI $\mathcal{M}^c_{\bm{x}}$ excluding the masked region specified by \texttt{mask}.
\end{itemize}
The $p$-value calculated by over-conditioning can be obtained by specifying the optional argument \texttt{inference\_mode} as \texttt{"over\_conditioning"}.
\begin{CodeChunk}
  \begin{CodeInput}
  >>> result = si_model.inference(
  ...     input=x, var=1.0, inference_mode="over_conditioning"
  ... )
  >>> print(result.p_value)
  \end{CodeInput}
\end{CodeChunk}
\begin{CodeChunk}
  \begin{CodeOutput}
  0.35340273690957236
  \end{CodeOutput}
\end{CodeChunk}
The naive $p$-value can be obtained using the \texttt{naive\_p\_value()} method.
\begin{CodeChunk}
  \begin{CodeInput}
  >>> print(result.naive_p_value())
  \end{CodeInput}
\end{CodeChunk}
\begin{CodeChunk}
  \begin{CodeOutput}
  0.092769583177523
  \end{CodeOutput}
\end{CodeChunk}
The Bonferroni-corrected $p$-value can be obtained using the \texttt{bonferroni\_p\_value()} method.
The Bonferroni-corrected $p$-value is computed using \texttt{log\_num\_comparisons}, which specifies the logarithm of the total hypotheses.
For a $16 \times 16$ image, we test two hypotheses per pixel, yielding $2^{256}$ total hypotheses:
\begin{CodeChunk}
  \begin{CodeInput}
  >>> print(
          result.bonferroni_p_value(log_num_comparisons=256 * np.log(2))
      )
  \end{CodeInput}
\end{CodeChunk}
\begin{CodeChunk}
  \begin{CodeOutput}
  1.0
  \end{CodeOutput}
\end{CodeChunk}
While the Bonferroni correction can controls the type I error rate, it is overly conservative and typically produces inflated $p$-values.
\subsection{Preset Hypotheses}
\label{subsec:preset_hypotheses}
The \texttt{si4onnx} provides three preset hypotheses that can perform three different tests:
\begin{itemize}
  \item \texttt{BackMeanDiff}:
  Tests the equality of mean values between the ROI and its complement.
  This corresponds to the Example B1 in Section~\ref{sec:problem_settings}.
  \item \texttt{NeighborMeanDiff}: 
  Tests the equality of mean values between the ROI and its neighborhood regions.
  This corresponds to the Example B2 in Section~\ref{sec:problem_settings}.
  \item \texttt{ReferenceMeanDiff}: 
  Tests whether the mean value difference between the observed image and the reference image in the ROI is equal.
  This corresponds to the Example B3 in Section~\ref{sec:problem_settings}.
\end{itemize}
These \texttt{PresetHypothesis} classes accepts the following common arguments:
\begin{itemize}
  \item \texttt{threshold}: 
  A threshold $\tau$ for defining the ROI $\mathcal{M}_{\bm{X}} := \{i \in [n]  \mid \cS(\bm{X}) \geq \tau\}$.
  \item \texttt{i\_idx}:
  An integer indicating input index of the ONNX model used for the hypothesis testing. Defaults to \texttt{0}.
  \item \texttt{o\_idx}: 
  An integer indicating output index of the ONNX model used for the hypothesis testing. Defaults to \texttt{0}.
  \item \texttt{post\_process}: 
  The list indicating the post-processing to be applied to the DL model's output used for the hypothesis testing.
  Post-processing operations can be specified as a list and are applied sequentially in the order specified.
  All these operations are linear or piecewise-linear and preserve the piecewise-linearity of the algorithm $\cA$, ensuring that the auto-conditioning remain applicable.
  The following post-processing operations are available:
  \begin{itemize}
    \item \texttt{InputDiff}:
    A class indicating the subtraction operation takes the input specified by \texttt{i\_idx} from the output specified by \texttt{o\_idx}.
    \item \texttt{Abs}: 
    A class indicating the operation takes the absolute value of the output specified by \texttt{o\_idx}.
    \item \texttt{Neq}:
    A class indicating the operation takes the negative value of the output specified by \texttt{o\_idx}.
    \item \texttt{AverageFilter}:
    A class indicating the operation applies an average filter to the output specified by \texttt{o\_idx}.
    The kernel size can be specified using the optional argument \texttt{kernel\_size}.
    \item \texttt{GaussianFilter}:
    A class indicating the operation applies a Gaussian filter to the output specified by \texttt{o\_idx}.
    The kernel size and standard deviation can be specified using the optional arguments \texttt{kernel\_size} and \texttt{sigma}.
    Defaults to \texttt{kernel\_size=3} and \texttt{sigma=1.0}.
  \end{itemize}
  \item \texttt{use\_norm}:
  A boolean indicating whether to normalize $\cS(\bm{X})$ before thresholding.
  Note that this is applied after the processing selected by \texttt{post\_process}.
  Defaults to \texttt{False}.
\end{itemize}
In addition to these common arguments, The class \texttt{NeighborMeanDiff} has the argument \texttt{neighborhood\_range} to specify the radius $r$ of the neighborhood region.
Also, users can define their own hypotheses and perform hypothesis testing by implementing a subclass that inherits from the \texttt{Hypothesis} class.
\subsection{Simulation Example}
\label{subsec:simulation}
We demonstrate the $p$-value properties of the three preset hypotheses using synthetic data generated from the Gaussian distribution $\mathcal{N}(\bm{0}, I)$.
\begin{itemize}
  \item VAE (\texttt{BackMeanDiff}): 
  We use the VAE for anomaly detection.
  We perform statistical testing to evaluate whether there is a significant difference between the mean intensity values of the anomalous region with a large reconstruction error and the other regions.
  This corresponds to the Example A1 and Example B1 in Section~\ref{sec:problem_settings}.
  \item U-Net (\texttt{NeighborMeanDiff}): 
  We use the U-Net for segmentation tasks.
  we perform statistical testing to evaluate whether there is a significant difference between the mean intensity values of regions identified as anomalous and their neighborhood regions.
  This corresponds to the Example A2 and Example B2 in Section~\ref{sec:problem_settings}.
  \item CAM (\texttt{ReferenceMeanDiff}): 
  We use the CAM for classification tasks.
  We perform statistical testing to evaluate whether there is a significant difference between the mean intensity values of the observed image and the reference image in the activated region.
  This corresponds to the Example A3 and Example B3 in Section~\ref{sec:problem_settings}.
\end{itemize}
We compare the $p$-values computed by the three methods:
\begin{itemize}
  \item \texttt{selective}:
  The selective $p$-values computed by the parametric programming.
  \item \texttt{bonferroni}: 
  The Bonferroni-corrected $p$-values.
  \item \texttt{naive}:
  The naive $p$-values computed by the naive method.
\end{itemize}
\paragraph{Simulations under the null hypothesis.}
We confirm the uniformity of the $p$-value for data following the null hypothesis.
The synthetic data following the null hypothesis is generated from $\bm{X} = (X_1, X_2, \ldots, X_{256})^\top \sim \mathcal{N}(\bm{0}, I)$.
We run 500 iterations for each model.
We generate data following the null hypothesis as follows:
\begin{CodeChunk}
  \begin{CodeInput}
  >>> null_data = si4onnx.data.SyntheticDataset(
  ...     size=500, shape=(1, 16, 16), local_signal=0, seed=0
  ... )
  >>> null_dataloader = DataLoader(null_data, batch_size=1)
  \end{CodeInput}
\end{CodeChunk}

In the VAE framework, the post-processing procedure involves computing the absolute difference between the input and output images to identify anomalous regions through reconstruction error.
This approach enables precise localization of anomalies by analyzing pixel-wise reconstruction discrepancies.
Furthermore, a Gaussian filter is applied to smooth the reconstruction error, effectively suppressing minor discrepancies while preserving the structural characteristics of anomalies.
The instance of VAE (\texttt{BackMeanDiff}) is created as follows:
\begin{CodeChunk}
  \begin{CodeInput}
    >>> si_model = si4onnx.load(
    ...     model=onnx.load("vae_16.onnx"),
    ...     hypothesis=si4onnx.hypothesis.BackMeanDiff(
    ...         threshold=1.0,
    ...         post_process=[
    ...             si4onnx.operators.InputDiff(),
    ...             si4onnx.operators.Abs(),
    ...             si4onnx.operators.GaussianFilter()
    ...         ]
    ...     )
    ... )
  \end{CodeInput}
\end{CodeChunk}

\begin{figure}[H]
  \centering
  \begin{minipage}{0.32\textwidth}
    \centering
    \includegraphics[width=1.0\textwidth]{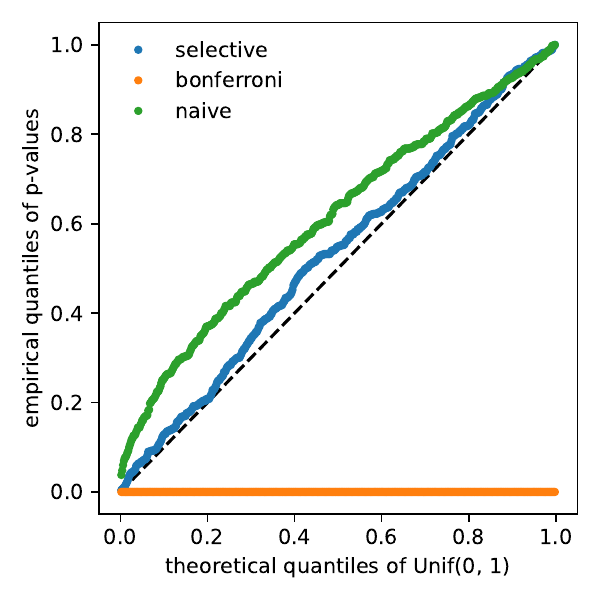}
    \subcaption{VAE}
  \end{minipage}
  \begin{minipage}{0.32\textwidth}
    \centering
    \includegraphics[width=1.0\textwidth]{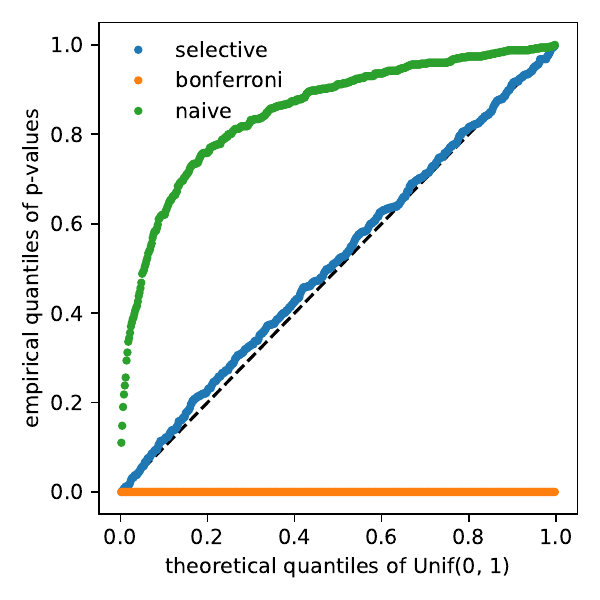}
    \subcaption{U-Net}
  \end{minipage}
  \begin{minipage}{0.32\textwidth}
    \centering
    \includegraphics[width=1.0\textwidth]{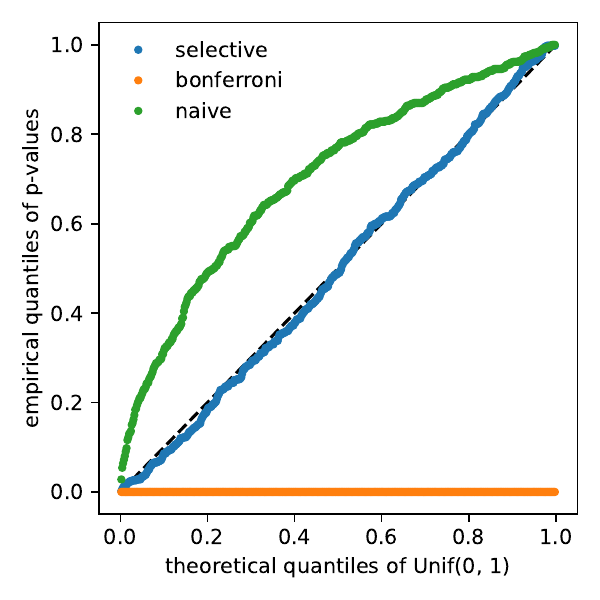}
    \subcaption{CAM}
  \end{minipage}
  \caption{Q-Q Plot for the synthetic data following the null hypothesis. The \texttt{selective} $p$-values exhibit uniform distribution under the null hypothesis, while other $p$-values deviate from uniformity.}
  \label{fig:qqplot_null}
\end{figure}

\begin{figure}[H]
  \centering
  \begin{minipage}{0.32\textwidth}
    \centering
    \includegraphics[width=1.0\textwidth]{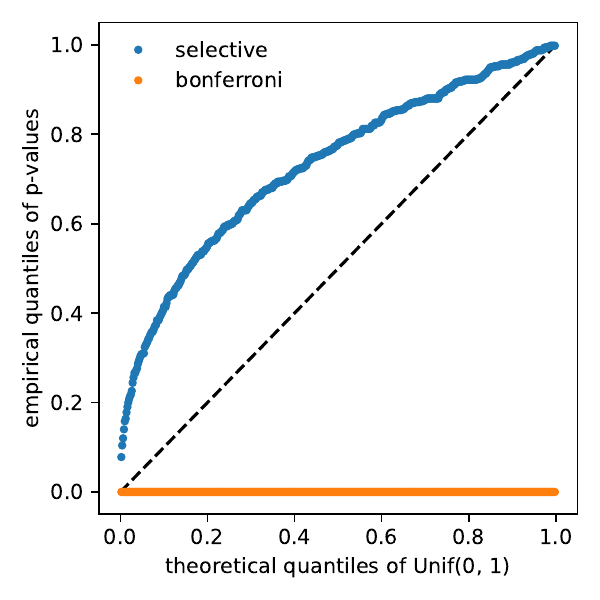}
    \subcaption{VAE}
  \end{minipage}
  \begin{minipage}{0.32\textwidth}
    \centering
    \includegraphics[width=1.0\textwidth]{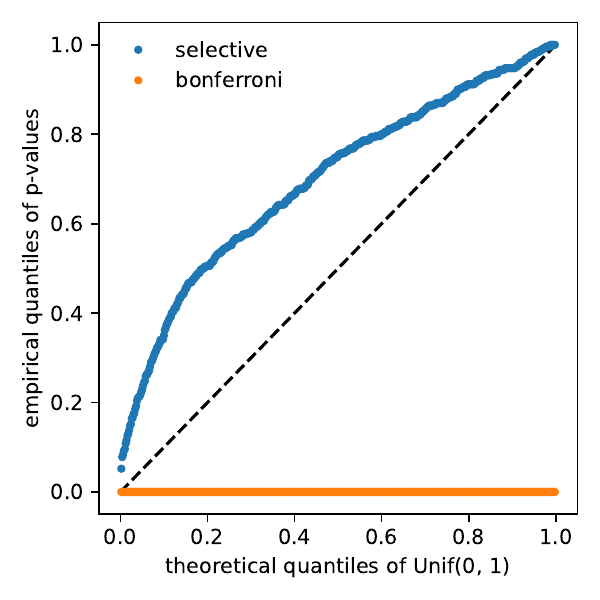}
    \subcaption{U-Net}
  \end{minipage}
  \begin{minipage}{0.32\textwidth}
    \centering
    \includegraphics[width=1.0\textwidth]{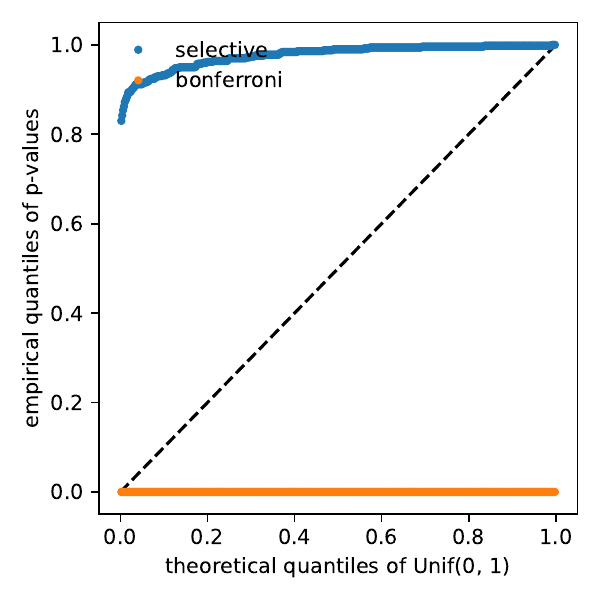}
    \subcaption{CAM}
  \end{minipage}
  \caption{Q-Q Plot for the synthetic data following the alternative hypothesis. The \texttt{selective} $p$-values demonstrate high power, while the \texttt{bonferroni} $p$-values show overly conservative behavior.}
  \label{fig:qqplot_alt}
\end{figure}

The U-Net architecture serves as a segmentation model in this simulation.
The model's output layer incorporates a sigmoid function.
Although the sigmoid function is nonlinear, \texttt{si4onnx} enables interval computation through automatic logit transformation of thresholds, even for outputs specified by the optional argument \texttt{o\_idx} that undergo sigmoid transformation.
The instance of U-Net (\texttt{NeighborMeanDiff}) is created as follows:
\begin{CodeChunk}
  \begin{CodeInput}
    >>> si_model = si4onnx.load(
    ...     model=onnx.load("unet_16.onnx"),
    ...     hypothesis=si4onnx.hypothesis.NeighborMeanDiff(
    ...         threshold=0.5,
    ...         neighborhood_range=1,
    ...     )
    ... )
  \end{CodeInput}
\end{CodeChunk}
The $p$-value can be computed for these two models as follows:
\begin{CodeChunk}
  \begin{CodeInput}
    >>> selective_p, naive_p, bonferroni_p = [], [], []
    >>> for x, _, _ in null_data:
    ...   result = si_model.inference(x, var=1.0)
    ...   selective_p.append(result.p_value)
    ...   naive_p.append(result.naive_p_value())
    ...   bonferroni_p.append(
    ...       result.bonferroni_p_value(log_num_comparisons=256 * np.log(2))
    ...   )
  \end{CodeInput}
\end{CodeChunk}
For the CAM (\texttt{ReferenceMeanDiff}), we first generate reference images following the null hypothesis:
\begin{CodeChunk}
  \begin{CodeInput}
    >>> ref_data = si4onnx.data.SyntheticDataset(
    ...     size=500, shape=(1, 16, 16), local_signal=0, seed=1
    ... )
    >> ref_dataloader = DataLoader(ref_data, batch_size=1)
  \end{CodeInput}
\end{CodeChunk}
Then, we create an instance of CAM with \texttt{ReferenceMeanDiff} hypothesis:
\begin{CodeChunk}
  \begin{CodeInput}
  >>> si_model = si4onnx.load(
  ...     model=onnx.load("cam_16.onnx"),
  ...     hypothesis=si4onnx.hypothesis.ReferenceMeanDiff(
  ...         threshold=0.8,
  ...         use_norm=True,
  ...     ),
  ... )
  \end{CodeInput}
\end{CodeChunk}
The \texttt{inference} method requires both a observed image and a reference image as input, provided as a tuple \texttt{(x, ref\_x)}.
For reference images, we use samples generated by the \texttt{SyntheticDataset} class.
To ensure independent $p$-values in this simulation, we use a different reference image for each observed image.
The following code demonstrates how to compute $p$-values for the CAM (\texttt{ReferenceMeanDiff}):
\begin{CodeChunk}
  \begin{CodeInput}
  >>> selective_p, naive_p, bonferroni_p = [], [], []
  >>> for x, _, _, ref_x, _, _ in zip(null_dataloader, ref_dataloader):
  ...   result = vae_si.inference((x, ref_x), var=1.0)
  ...   selective_p.append(result.p_value)
  ...   naive_p.append(result.naive_p_value())
  ...   bonferroni_p.append(
  ...       result.bonferroni_p_value(log_num_comparisons=256 * np.log(2))
  ...   )
  \end{CodeInput}
\end{CodeChunk}
Fig.~\ref{fig:qqplot_null} shows the Q-Q plots of these $p$-values.
The \texttt{naive} $p$-values do not follow a uniform distribution, indicating that the type I error rate exceeds the significance level.
The \texttt{bonferroni} $p$-values are very conservative.
On the other hand, the \texttt{selective} $p$-values follow a uniform distribution, indicating that the test is valid.
\paragraph{Simulations under the alternative hypothesis.}
We confirm the $p$-values for data following the alternative hypothesis.
The synthetic data following the alternative hypothesis is generated from $\bm{X} = (X_1, X_2, \ldots, X_{256})^\top \sim \mathcal{N}(\bm{\mu}, I)$, where the mean vector $\bm{\mu}$ is set to $\mu_i = 3$ for $i \in \mathcal{S}$ and $\mu_i = 0$ for $i \in [n] \backslash \mathcal{S}$.
The synthetic data following the alternative hypothesis is generated as follows:
\begin{CodeChunk}
  \begin{CodeInput}
  >>> alt_data = si4onnx.data.SyntheticDataset(
  ...     size=500, shape=(1, 16, 16), local_signal=3, seed=0
  ... )
  >>> alt_dataloader = DataLoader(alt_data, batch_size=1)
  \end{CodeInput}
\end{CodeChunk}
The inference can be performed in the same way as for the null hypothesis, replacing the instance \texttt{null\_dataloader} with \texttt{alt\_dataloader}.
Fig.~\ref{fig:qqplot_alt} shows the Q-Q plots for data following the alternative hypothesis.
The \texttt{naive} $p$-values are not considered because they do not follow a uniform distribution under the null hypothesis.
The \texttt{bonferroni} $p$-values indicating low power, as they are very conservative.
The \texttt{selective} $p$-values show high power.

\newpage
\section{Discussion and Conclusion}
\label{sec:discussion}
Evaluating the statistical significance of ROIs obtained from deep learning models is crucial to assessing AI reliability, with selective inference serving as an effective method for quantitative reliability evaluation.
The Python package \texttt{si4onnx} facilitates selective inference implementation for deep learning models developed across various frameworks.
While computational complexity theoretically increases exponentially with the number of neurons in the model's piecewise linear functions, practical computational costs inevitably rise, though not necessarily exponentially.
Although \texttt{si4onnx} employs memoization to reduce computational costs, further performance optimization can be achieved through ONNX model optimization via the ONNX ecosystem and parallel computing capabilities provided by \texttt{sicore}.
In the future, we plan to improve the computational time and add more diverse preset hypotheses and post-processing, as well as add support for additional layers.

\newpage
\subsection*{Acknowledgement}
This work was partially supported by MEXT KAKENHI (20H00601), JST CREST (JPMJCR21D3, JPMJCR22N2), JST Moonshot R\&D (JPMJMS2033-05), JST AIP Acceleration Research (JPMJCR21U2), NEDO (JPNP18002, JPNP20006) and RIKEN Center for Advanced Intelligence Project.

\clearpage
%


%
\newpage
\bibliographystyle{plainnat}
\bibliography{ref}

\begin{thebibliography}{40}
\providecommand{\natexlab}[1]{#1}
\providecommand{\url}[1]{\texttt{#1}}
\expandafter\ifx\csname urlstyle\endcsname\relax
  \providecommand{\doi}[1]{doi: #1}\else
  \providecommand{\doi}{doi: \begingroup \urlstyle{rm}\Url}\fi

\bibitem[Adebayo et~al.(2018)Adebayo, Gilmer, Muelly, Goodfellow, Hardt, and
  Kim]{adebayo2018sanity}
Julius Adebayo, Justin Gilmer, Michael Muelly, Ian Goodfellow, Moritz Hardt,
  and Been Kim.
\newblock Sanity checks for saliency maps.
\newblock \emph{Advances in neural information processing systems}, 31, 2018.

\bibitem[Bergmann et~al.(2019)Bergmann, Fauser, Sattlegger, and
  Steger]{bergmann2019mvtec}
Paul Bergmann, Michael Fauser, David Sattlegger, and Carsten Steger.
\newblock Mvtec ad--a comprehensive real-world dataset for unsupervised anomaly
  detection.
\newblock In \emph{Proceedings of the IEEE/CVF conference on computer vision
  and pattern recognition}, pages 9592--9600, 2019.

\bibitem[Chen and Bien(2020)]{chen2020valid}
Shuxiao Chen and Jacob Bien.
\newblock Valid inference corrected for outlier removal.
\newblock \emph{Journal of Computational and Graphical Statistics}, 29\penalty0
  (2):\penalty0 323--334, 2020.

\bibitem[Chen et~al.(2023)Chen, Jewell, and Witten]{chen2023more}
Yiqun Chen, Sean Jewell, and Daniela Witten.
\newblock More powerful selective inference for the graph fused lasso.
\newblock \emph{Journal of Computational and Graphical Statistics}, 32\penalty0
  (2):\penalty0 577--587, 2023.

\bibitem[Das et~al.(2021)Das, Duy, Hanada, Tsuda, and Takeuchi]{das2021fast}
Diptesh Das, Vo~Nguyen~Le Duy, Hiroyuki Hanada, Koji Tsuda, and Ichiro
  Takeuchi.
\newblock Fast and more powerful selective inference for sparse high-order
  interaction model.
\newblock \emph{arXiv preprint arXiv:2106.04929}, 2021.

\bibitem[Dombrowski et~al.(2019)Dombrowski, Alber, Anders, Ackermann,
  M{\"u}ller, and Kessel]{dombrowski2019explanations}
Ann-Kathrin Dombrowski, Maximillian Alber, Christopher Anders, Marcel
  Ackermann, Klaus-Robert M{\"u}ller, and Pan Kessel.
\newblock Explanations can be manipulated and geometry is to blame.
\newblock \emph{Advances in neural information processing systems}, 32, 2019.

\bibitem[Duy and Takeuchi(2022)]{duy2022more}
Vo~Nguyen~Le Duy and Ichiro Takeuchi.
\newblock More powerful conditional selective inference for generalized lasso
  by parametric programming.
\newblock \emph{The Journal of Machine Learning Research}, 23\penalty0
  (1):\penalty0 13544--13580, 2022.

\bibitem[Duy et~al.(2020)Duy, Toda, Sugiyama, and Takeuchi]{duy2020computing}
Vo~Nguyen~Le Duy, Hiroki Toda, Ryota Sugiyama, and Ichiro Takeuchi.
\newblock Computing valid p-value for optimal changepoint by selective
  inference using dynamic programming.
\newblock In \emph{Advances in Neural Information Processing Systems}, 2020.

\bibitem[Duy et~al.(2022)Duy, Iwazaki, and Takeuchi]{duy2022quantifying}
Vo~Nguyen~Le Duy, Shogo Iwazaki, and Ichiro Takeuchi.
\newblock Quantifying statistical significance of neural network-based image
  segmentation by selective inference.
\newblock \emph{Advances in Neural Information Processing Systems},
  35:\penalty0 31627--31639, 2022.

\bibitem[Gao et~al.(2022)Gao, Bien, and Witten]{gao2022selective}
Lucy~L Gao, Jacob Bien, and Daniela Witten.
\newblock Selective inference for hierarchical clustering.
\newblock \emph{Journal of the American Statistical Association}, pages 1--11,
  2022.

\bibitem[Garcia-Angulo and Claeskens(2023)]{garcia2023optimal}
Andrea~C Garcia-Angulo and Gerda Claeskens.
\newblock Optimal finite sample post-selection confidence distributions in
  generalized linear models.
\newblock \emph{Journal of Statistical Planning and Inference}, 222:\penalty0
  66--77, 2023.

\bibitem[Golan and El-Yaniv(2018)]{golan2018deep}
Izhak Golan and Ran El-Yaniv.
\newblock Deep anomaly detection using geometric transformations.
\newblock \emph{Advances in neural information processing systems}, 31, 2018.

\bibitem[Hyun et~al.(2018)Hyun, G{'}sell, and Tibshirani]{hyun2018exact}
Sangwon Hyun, Max G{'}sell, and Ryan~J Tibshirani.
\newblock Exact post-selection inference for the generalized lasso path.
\newblock \emph{Electronic Journal of Statistics}, 12\penalty0 (1):\penalty0
  1053--1097, 2018.

\bibitem[Jewell et~al.(2022)Jewell, Fearnhead, and Witten]{jewell2022testing}
Sean Jewell, Paul Fearnhead, and Daniela Witten.
\newblock Testing for a change in mean after changepoint detection.
\newblock \emph{Journal of the Royal Statistical Society Series B: Statistical
  Methodology}, 84\penalty0 (4):\penalty0 1082--1104, 2022.

\bibitem[Karargyris et~al.(2023)Karargyris, Umeton, Sheller, Aristizabal,
  George, Wuest, Pati, Kassem, Zenk, Baid, {Narayana Moorthy}, Chowdhury, Guo,
  Nalawade, Rosenthal, Kanter, Xenochristou, Beutel, Chung, Bergquist, Eddy,
  Abid, Tunstall, Sanseviero, Dimitriadis, Qian, Xu, Liu, Goh, Bala, Bittorf,
  Puchala, Ricciuti, Samineni, Sengupta, Chaudhari, Coleman, Desinghu, Diamos,
  Dutta, Feddema, Fursin, Huang, Kashyap, Lane, Mallick, Mascagni, Mehta,
  Moraes, Natarajan, Nikolov, Padoy, Pekhimenko, Reddi, Reina, Ribalta, Singh,
  Thiagarajan, Albrecht, Wolf, Miller, Fu, Shah, Xu, Yadav, Talby, Awad,
  Howard, Rosenthal, Marchionni, Loda, Johnson, Bakas, Mattson, {FeTS
  Consortium}, {BraTS-2020 Consortium}, and {AI4SafeChole
  Consortium}]{Karargyris2023}
Alexandros Karargyris, Renato Umeton, Micah~J. Sheller, Alejandro Aristizabal,
  Johnu George, Anna Wuest, Sarthak Pati, Hasan Kassem, Maximilian Zenk, Ujjwal
  Baid, Prakash {Narayana Moorthy}, Alexander Chowdhury, Junyi Guo, Sahil
  Nalawade, Jacob Rosenthal, David Kanter, Maria Xenochristou, Daniel~J.
  Beutel, Verena Chung, Timothy Bergquist, James Eddy, Abubakar Abid, Lewis
  Tunstall, Omar Sanseviero, Dimitrios Dimitriadis, Yiming Qian, Xinxing Xu,
  Yong Liu, Rick Siow~Mong Goh, Srini Bala, Victor Bittorf, Sreekar~Reddy
  Puchala, Biagio Ricciuti, Soujanya Samineni, Eshna Sengupta, Akshay
  Chaudhari, Cody Coleman, Bala Desinghu, Gregory Diamos, Debo Dutta, Diane
  Feddema, Grigori Fursin, Xinyuan Huang, Satyananda Kashyap, Nicholas Lane,
  Indranil Mallick, Pietro Mascagni, Virendra Mehta, Cassiano~Ferro Moraes,
  Vivek Natarajan, Nikola Nikolov, Nicolas Padoy, Gennady Pekhimenko,
  Vijay~Janapa Reddi, G.~Anthony Reina, Pablo Ribalta, Abhishek Singh,
  Jayaraman~J. Thiagarajan, Jacob Albrecht, Thomas Wolf, Geralyn Miller, Huazhu
  Fu, Prashant Shah, Daguang Xu, Poonam Yadav, David Talby, Mark~M. Awad,
  Jeremy~P. Howard, Michael Rosenthal, Luigi Marchionni, Massimo Loda, Jason~M.
  Johnson, Spyridon Bakas, Peter Mattson, {FeTS Consortium}, {BraTS-2020
  Consortium}, and {AI4SafeChole Consortium}.
\newblock Federated benchmarking of medical artificial intelligence with
  medperf.
\newblock \emph{Nature Machine Intelligence}, 5\penalty0 (7):\penalty0
  799--810, July 2023.
\newblock \doi{10.1038/s42256-023-00652-2}.
\newblock URL \url{https://doi.org/10.1038/s42256-023-00652-2}.

\bibitem[Katsuoka et~al.(2024)Katsuoka, Shiraishi, Miwa, Duy, and
  Takeuchi]{katsuoka2024statisticaltestdiffusionmodelbased}
Teruyuki Katsuoka, Tomohiro Shiraishi, Daiki Miwa, Vo~Nguyen~Le Duy, and Ichiro
  Takeuchi.
\newblock Statistical test on diffusion model-based generated images by
  selective inference, 2024.
\newblock URL \url{https://arxiv.org/abs/2402.11789}.

\bibitem[Kriegeskorte et~al.(2009)Kriegeskorte, Simmons, Bellgowan, and
  Baker]{kriegeskorte2009circular}
Nikolaus Kriegeskorte, W~Kyle Simmons, Patrick~SF Bellgowan, and Chris~I Baker.
\newblock Circular analysis in systems neuroscience: the dangers of double
  dipping.
\newblock \emph{Nature neuroscience}, 12\penalty0 (5):\penalty0 535--540, 2009.

\bibitem[LaBella et~al.(2023)LaBella, Adewole, Alonso-Basanta, Altes, Anwar,
  Baid, Bergquist, Bhalerao, Chen, Chung, Conte, Dako, Eddy, Ezhov, Godfrey,
  Hilal, Familiar, Farahani, Iglesias, Jiang, Johanson, Kazerooni, Kent,
  Kirkpatrick, Kofler, Leemput, Li, Liu, Mahtabfar, McBurney-Lin, McLean,
  Meier, Moawad, Mongan, Nedelec, Pajot, Piraud, Rashid, Reitman, Shinohara,
  Velichko, Wang, Warman, Wiggins, Aboian, Albrecht, Anazodo, Bakas, Flanders,
  Janas, Khanna, Linguraru, Menze, Nada, Rauschecker, Rudie, Tahon,
  Villanueva-Meyer, Wiestler, and
  Calabrese]{labella2023asnrmiccaibraintumorsegmentation}
Dominic LaBella, Maruf Adewole, Michelle Alonso-Basanta, Talissa Altes,
  Syed~Muhammad Anwar, Ujjwal Baid, Timothy Bergquist, Radhika Bhalerao, Sully
  Chen, Verena Chung, Gian-Marco Conte, Farouk Dako, James Eddy, Ivan Ezhov,
  Devon Godfrey, Fathi Hilal, Ariana Familiar, Keyvan Farahani, Juan~Eugenio
  Iglesias, Zhifan Jiang, Elaine Johanson, Anahita~Fathi Kazerooni, Collin
  Kent, John Kirkpatrick, Florian Kofler, Koen~Van Leemput, Hongwei~Bran Li,
  Xinyang Liu, Aria Mahtabfar, Shan McBurney-Lin, Ryan McLean, Zeke Meier,
  Ahmed~W Moawad, John Mongan, Pierre Nedelec, Maxence Pajot, Marie Piraud,
  Arif Rashid, Zachary Reitman, Russell~Takeshi Shinohara, Yury Velichko,
  Chunhao Wang, Pranav Warman, Walter Wiggins, Mariam Aboian, Jake Albrecht,
  Udunna Anazodo, Spyridon Bakas, Adam Flanders, Anastasia Janas, Goldey
  Khanna, Marius~George Linguraru, Bjoern Menze, Ayman Nada, Andreas~M
  Rauschecker, Jeff Rudie, Nourel~Hoda Tahon, Javier Villanueva-Meyer, Benedikt
  Wiestler, and Evan Calabrese.
\newblock The asnr-miccai brain tumor segmentation (brats) challenge 2023:
  Intracranial meningioma, 2023.
\newblock URL \url{https://arxiv.org/abs/2305.07642}.

\bibitem[Le~Duy et~al.(2024)Le~Duy, Lin, and Takeuchi]{le2024cad}
Vo~Nguyen Le~Duy, Hsuan-Tien Lin, and Ichiro Takeuchi.
\newblock Cad-da: Controllable anomaly detection after domain adaptation by
  statistical inference.
\newblock In \emph{International Conference on Artificial Intelligence and
  Statistics}, pages 1828--1836. PMLR, 2024.

\bibitem[Lee and Taylor(2014)]{lee2014exact}
Jason~D Lee and Jonathan~E Taylor.
\newblock Exact post model selection inference for marginal screening.
\newblock \emph{Advances in neural information processing systems}, 27, 2014.

\bibitem[Lee et~al.(2015)Lee, Sun, and Taylor]{lee2015evaluating}
Jason~D Lee, Yuekai Sun, and Jonathan~E Taylor.
\newblock Evaluating the statistical significance of biclusters.
\newblock \emph{Advances in neural information processing systems}, 28, 2015.

\bibitem[Lee et~al.(2016)Lee, Sun, Sun, and Taylor]{lee2016exact}
Jason~D Lee, Dennis~L Sun, Yuekai Sun, and Jonathan~E Taylor.
\newblock Exact post-selection inference, with application to the lasso.
\newblock \emph{The Annals of Statistics}, 44\penalty0 (3):\penalty0 907--927,
  2016.

\bibitem[Liu et~al.(2018)Liu, Markovic, and Tibshirani]{liu2018more}
Keli Liu, Jelena Markovic, and Robert Tibshirani.
\newblock More powerful post-selection inference, with application to the
  lasso.
\newblock \emph{arXiv preprint arXiv:1801.09037}, 2018.

\bibitem[Long et~al.(2015)Long, Shelhamer, and Darrell]{long2015fully}
Jonathan Long, Evan Shelhamer, and Trevor Darrell.
\newblock Fully convolutional networks for semantic segmentation.
\newblock In \emph{Proceedings of the IEEE conference on computer vision and
  pattern recognition}, pages 3431--3440, 2015.

\bibitem[Miwa et~al.(2023)Miwa, Le, and Takeuchi]{miwa2023valid}
Daiki Miwa, Duy Vo~Nguyen Le, and Ichiro Takeuchi.
\newblock Valid p-value for deep learning-driven salient region.
\newblock In \emph{Proceedings of the 11th International Conference on Learning
  Representation}, 2023.

\bibitem[Miwa et~al.(2024)Miwa, Shiraishi, Duy, Katsuoka, and
  Takeuchi]{miwa2024statistical}
Daiki Miwa, Tomohiro Shiraishi, Vo~Nguyen~Le Duy, Teruyuki Katsuoka, and Ichiro
  Takeuchi.
\newblock Statistical test for anomaly detections by variational auto-encoders.
\newblock \emph{arXiv preprint arXiv:2402.03724}, 2024.

\bibitem[Pirenne and Claeskens(2024)]{pirenne2024parametric}
Sarah Pirenne and Gerda Claeskens.
\newblock Parametric programming-based approximate selective inference for
  adaptive lasso, adaptive elastic net and group lasso.
\newblock \emph{Journal of Statistical Computation and Simulation}, pages
  1--24, 2024.

\bibitem[Ronneberger et~al.(2015)Ronneberger, Fischer, and
  Brox]{ronneberger2015u}
Olaf Ronneberger, Philipp Fischer, and Thomas Brox.
\newblock U-net: Convolutional networks for biomedical image segmentation.
\newblock In \emph{Medical image computing and computer-assisted
  intervention--MICCAI 2015: 18th international conference, Munich, Germany,
  October 5-9, 2015, proceedings, part III 18}, pages 234--241. Springer, 2015.

\bibitem[R{\"u}gamer and Greven(2020)]{rugamer2020inference}
David R{\"u}gamer and Sonja Greven.
\newblock Inference for l 2-boosting.
\newblock \emph{Statistics and computing}, 30\penalty0 (2):\penalty0 279--289,
  2020.

\bibitem[Selvaraju et~al.(2017)Selvaraju, Cogswell, Das, Vedantam, Parikh, and
  Batra]{selvaraju2017grad}
Ramprasaath~R Selvaraju, Michael Cogswell, Abhishek Das, Ramakrishna Vedantam,
  Devi Parikh, and Dhruv Batra.
\newblock Grad-cam: Visual explanations from deep networks via gradient-based
  localization.
\newblock In \emph{Proceedings of the IEEE international conference on computer
  vision}, pages 618--626, 2017.

\bibitem[Shiraishi et~al.(2024{\natexlab{a}})Shiraishi, Miwa, Duy, and
  Takeuchi]{Shiraishi2024}
Tomohiro Shiraishi, Daiki Miwa, Vo~Nguyen~Le Duy, and Ichiro Takeuchi.
\newblock Bounded p values in parametric programming-based selective inference.
\newblock \emph{Japanese Journal of Statistics and Data Science}, 04
  2024{\natexlab{a}}.
\newblock ISSN 2520-8764.
\newblock \doi{10.1007/s42081-024-00247-0}.
\newblock URL \url{https://doi.org/10.1007/s42081-024-00247-0}.

\bibitem[Shiraishi et~al.(2024{\natexlab{b}})Shiraishi, Miwa, Katsuoka, Duy,
  Taji, and Takeuchi]{shiraishi2024statistical}
Tomohiro Shiraishi, Daiki Miwa, Teruyuki Katsuoka, Vo~Nguyen~Le Duy, Koichi
  Taji, and Ichiro Takeuchi.
\newblock Statistical test for attention map in vision transformers.
\newblock \emph{International Conference on Machine Learning},
  2024{\natexlab{b}}.

\bibitem[Suzumura et~al.(2017)Suzumura, Nakagawa, Umezu, Tsuda, and
  Takeuchi]{suzumura2017selective}
Shinya Suzumura, Kazuya Nakagawa, Yuta Umezu, Koji Tsuda, and Ichiro Takeuchi.
\newblock Selective inference for sparse high-order interaction models.
\newblock In \emph{Proceedings of the 34th International Conference on Machine
  Learning-Volume 70}, pages 3338--3347. JMLR. org, 2017.

\bibitem[Tanizaki et~al.(2020)Tanizaki, Hashimoto, Inatsu, Hontani, and
  Takeuchi]{tanizaki2020computing}
Kosuke Tanizaki, Noriaki Hashimoto, Yu~Inatsu, Hidekata Hontani, and Ichiro
  Takeuchi.
\newblock Computing valid p-values for image segmentation by selective
  inference.
\newblock In \emph{Proceedings of the IEEE/CVF Conference on Computer Vision
  and Pattern Recognition}, pages 9553--9562, 2020.

\bibitem[Taylor and Tibshirani(2015)]{taylor2015statistical}
Jonathan Taylor and Robert~J Tibshirani.
\newblock Statistical learning and selective inference.
\newblock \emph{Proceedings of the National Academy of Sciences}, 112\penalty0
  (25):\penalty0 7629--7634, 2015.

\bibitem[Tian and Taylor(2018)]{tian2018selective}
Xiaoying Tian and Jonathan Taylor.
\newblock Selective inference with a randomized response.
\newblock \emph{The Annals of Statistics}, 46\penalty0 (2):\penalty0 679--710,
  2018.

\bibitem[Tibshirani et~al.(2016)Tibshirani, Taylor, Lockhart, and
  Tibshirani]{tibshirani2016exact}
Ryan~J Tibshirani, Jonathan Taylor, Richard Lockhart, and Robert Tibshirani.
\newblock Exact post-selection inference for sequential regression procedures.
\newblock \emph{Journal of the American Statistical Association}, 111\penalty0
  (514):\penalty0 600--620, 2016.

\bibitem[Tsukurimichi et~al.(2021)Tsukurimichi, Inatsu, Duy, and
  Takeuchi]{tsukurimichi2021conditional}
Toshiaki Tsukurimichi, Yu~Inatsu, Vo~Nguyen~Le Duy, and Ichiro Takeuchi.
\newblock Conditional selective inference for robust regression and outlier
  detection using piecewise-linear homotopy continuation.
\newblock \emph{arXiv preprint arXiv:2104.10840}, 2021.

\bibitem[Yang et~al.(2016)Yang, Barber, Jain, and Lafferty]{yang2016selective}
Fan Yang, Rina~Foygel Barber, Prateek Jain, and John Lafferty.
\newblock Selective inference for group-sparse linear models.
\newblock In \emph{Advances in Neural Information Processing Systems}, pages
  2469--2477, 2016.

\bibitem[Zhou et~al.(2016)Zhou, Khosla, Lapedriza, Oliva, and
  Torralba]{zhou2016learning}
Bolei Zhou, Aditya Khosla, Agata Lapedriza, Aude Oliva, and Antonio Torralba.
\newblock Learning deep features for discriminative localization.
\newblock In \emph{Proceedings of the IEEE conference on computer vision and
  pattern recognition}, pages 2921--2929, 2016.

\end{thebibliography}

\end{document}